\pdfoutput=1

\documentclass[11pt]{article}

\usepackage{EMNLP2022}

\usepackage{times}
\usepackage{latexsym}
\usepackage{graphicx}
\usepackage[T1]{fontenc}

\usepackage[utf8]{inputenc}

\usepackage{microtype}

\usepackage{inconsolata}
\usepackage{hyperref}

\title{The VolcTrans System for WMT22 Multilingual Machine Translation Task}

\author{Xian Qian$^{1}$, Kai Hu$^{1}$, Jiaqiang Wang$^{1}$, Yifeng Liu$^{2}$ \\ {\bf Xingyuan Pan$^3$, Jun Cao$^{1}$, Mingxuan Wang$^{1}$} \\
  $^1$ ByteDance AI Lab,  
  $^2$ Tsinghua University, $^3$ Wuhan University \\
  \texttt{\{qian.xian, 
hukai.joseph,wangjiaqiang.sonian,}\\{caojun.sh, wangmingxuan.89\}@bytedance.com} \\ \texttt{liuyifen20@mails.tsinghua.edu.cn, panxingyuan209@gmail.com}
}

\begin{document}
\maketitle
\begin{abstract}
This report describes our VolcTrans system for the WMT22 shared task on large-scale multilingual machine translation. We participated in the unconstrained track which allows the use of external resources. Our system is a transformer-based multilingual model trained on data from multiple sources including the public training set from the data track, NLLB data provided by Meta AI, self-collected parallel corpora, and pseudo bitext from back-translation. A series of heuristic rules clean both bilingual and monolingual texts. On the official test set, our system achieves $17.3$ BLEU, $21.9$ spBLEU, and $41.9$ chrF2++ on average over all language pairs. The average inference speed is $11.5$ sentences per second using a single Nvidia Tesla V100 GPU. Our code and trained models
are available at \href{https://github.com/xian8/wmt22}{https://github.com/xian8/wmt22}
\end{abstract}

\section{Introduction}
Multilingual Machine Translation attracts much attention in recent years due to its advantages in sharing cross-lingual knowledge for low-resource languages. It also dramatically reduces training and serving costs. Training a multilingual model is much faster and simpler than training many bilingual ones. Serving multiple low-traffic languages using one model could drastically improve GPU utilization.

The WMT22 shared task on large-scale multilingual machine translation includes 24 African languages \cite{adelani-etal-2022-findings}.  Inspired by previous research works, we train a deep transformer model to translate all languages since large models have been demonstrated effective for multilingual translation \cite{JMLR:v22:20-1307,kong-etal-2021-multilingual,zhang-etal-2020-improving}. We participated in the unconstrained track that allows the use of external data. Besides the official dataset for the constrained track, and the NLLB corpus provided by MetaAI \cite{https://doi.org/10.48550/arxiv.2207.04672}, we also collect parallel and monolingual texts from public websites and sources. These raw data are cleaned by a series of commonly used heuristic rules, and a minimum description length (MDL) based approach to remove samples with repeat patterns. Monolingual texts are used for back translation. For some very low-resource languages such as Wolof, iterative back-translation is adopted for higher accuracy. 

We compare different training strategies to balance efficiency and quality, such as streaming data shuffling, and dynamic vocabulary for new languages. Furthermore, we used the open-sourced LightSeq toolkit \footnote{https://github.com/bytedance/lightseq} to accelerate training and inference. 

On the official test set, our system achieves  $17.3$ BLEU, $21.9$ spBLEU, and $41.9$ chrF2++ on average over all language pairs. Averaged inference speed is $11.5$ sentences per second using a single Nvidia Tesla V100 GPU.  

\section{Data}

\subsection{Data Collection}

Our training data are mainly from four sources: the official set for constrained track, NLLB data provided by Meta AI, self-collected corpora, and pseudo training set from back translation.

For each source, we collect both parallel sentence pairs and monolingual sentences.
A parallel sentence pair is collected if one side is in African language and the other is in English or French.
We did not collect African-African sentence pairs as we use English as the pivot language for African-to-African translation. Instead, they are added to the monolingual set. More specifically, we split every sentence pair into two sentences and add them to the monolingual set accordingly. For example, the source side of a fuv-fon sentence pair is added to the fuv set. This greatly enriches the monolingual dataset, especially for the very low-resource languages.

We merge multiple corpora from the same source into one and use bloom filter \footnote{https://pypi.org/project/bloom-filter}\cite{Bloom70space/timetrade-offs} for fast deduplication. To reduce false positive errors which over delete distinct samples, we set the error rate $1e-7$ and capacity of $4B$ samples which costs $100G$ host memory. 

The official set includes the data from data track participants, OPUS collections, and the NLLB parallel corpora mined from Common Crawl \cite{common-crawl} and other sources. All domains in OPUS collections are involved, such as Mozilla-I10n, which could introduce many noises such as programming languages, and needs extra rules to clean.

NLLB data provided by Meta AI has three subsets: primary bitext including a seed set that is carefully annotated for representative languages and a public bitext set downloaded from open sources and mined bitexts that are automatically discovered by LASER3 encoder in a global mining pipeline, back-translated data from a pretrained model. We add the first two subsets in our training set.

Some public bitext data that are no longer available or require authorization such as JW300 \cite{agic-vulic-2019-jw300}, Lorelei\footnote{https://catalog.ldc.upenn.edu/LDC2021T02} and Chichewa News \footnote{https://zenodo.org/record/4315018\#.YypJWezML0p} are not included. We noticed that the NLLB team released another version of mined data recently in hugging-face \footnote{https://huggingface.co/datasets/allenai/nllb}, which is different from the version on the WMT22 website. We merge the new version into the old one and remove duplicates.

We collected additional bitexts in two ways: large-scale mining from general web pages,  and manually crawling from specific websites and sources. 

Large-scale mining focused on two scenarios, parallel sentences appearing on a single web page such as dictionary web pages that use multiple bilingual sentences to exemplify the usage of a word, and parallel web pages that describe the same content but are written in different languages.  We extract these pages from the Common Crawl corpus. Then we utilized Vecalign \cite{thompson-koehn-2019-vecalign}, an accurate and efficient sentence alignment algorithm to mine parallel bilingual sentences. We use LASER \cite{DBLP:conf/rep4nlp/SchwenkD17} encoders released by WMT to obtain multilingual sentence embeddings and facilitate the alignment work. We collected about 3 million sentence pairs namely LAVA corpus and submitted them to the data track. And another $150M$ pairs for the unconstrained track.

Specific websites and sources have fewer but higher-quality sentence pairs. For example, the bible website\footnote{https://www.bible.com/languages} labels the order of sentences across languages so we can align them easily without sentence segmentation. Since JW300 is not publicly available, we crawled pages from Jehovah’s Witnesses\footnote{https://www.jw.org} to recover the dataset. 

Monolingual texts have richer sources such as VOA news in Amharic \footnote{https://amharic.voanews.com/} and OSCAR \cite{2022arXiv220106642A}, which improve English/French $\rightarrow $ African translation using back-translation. Monolingual texts from parallel data are also collected as described above. For African $\rightarrow$ English/French translation, we clean Wikipedia pages in English/French to get monolingual texts. For languages that gain significantly from back-translation such as Wolof, we run another round of back-translation to generate high-quality pseudo data.

\subsection{Data Cleaning}

We used the following rules to clean parallel datasets, except the NLLB mined bitext.

\begin{itemize}
    \item Filter out parentheses and texts in between if the numbers of parentheses in two sentences are different. 
    \item Filter out sentence pairs if numbers mismatch or one sentence ends with punctuation \textit{: ! ? ...} and the other mismatches.
    \item Filter out sentences shorter than 30 characters, sentences having URLs or emails, or words longer than 100 characters.
    \item De-duplication: remove sentence pairs sharing the same source or target but having different translations.
    \item Sentences having programming languages are removed. We manually create a set of keywords to detect programming languages, such as \textit{if (}\  , \textit{==} \  and \textit{.getAttribute} .
    \item Language identification using the NLLB language identification model trained by fastText \cite{joulin-etal-2017-bag}
\end{itemize}

One type of noisy text could survive the rules above, which has repeat patterns and commonly exists in many datasets. Here are some examples,
\noindent\fbox{%
    \parbox{\linewidth}{\begin{small}%
        \quad \textit{Download Bongeziwe Mabandla mini esadibana ngayo (\#001) Mp3 Bongeziwe Mabandla - mini esadibana ngayo (\#001).}

\quad \textit{Coaster Gift,Paper-Cut Coaster Zodiac,Red Coaster Cute,Paper-Cut Zodiac Coaster} 

\quad  \textit{mm mm mm MPEE(um) MPEP(um) mm mm mm mm mm mm kg kg}
    \end{small}
}}

A natural choice to detect these repeating patterns is the minimum description length (MDL) which finds the optimal compression by encoding frequent substrings with shorter codes.

Specifically, given a sentence $\mathbf{s}$, our MDL objective minimizes the length of the codebook plus the bits to encode the sentence:
\begin{eqnarray}
     \mathrm{MDL}(\mathbf{s}) &= &\min_{\mathbf{s}=  w_1w_2\dots w_n}  \left(C\sum_{\mathrm{distinct\ } w} |w|\right.\nonumber\\ && \left.-\sum_i \log\left(p(w_i|w_{i-1})\right) \right) \nonumber
\end{eqnarray}
where $w_1, w_2, \dots w_n$ is the word (coding entry) sequence,  $C$ is a positive constant, which balances the contribution of the codebook and length of the encoded sequence. $|w|$ is the length of word $w$. In our experiments, we set $C=2$. $p(w_i|w_{i-1}) = \frac{\#w_{i-1}w_i}{\#w_{i-1}}$ is the conditional probability of word bigrams in the sequence.

A sentence is noisy if the ratio of MDL over sentence length is less than a predefined threshold:
\begin{eqnarray}
\mathbf{s}\ \mathrm{is\ noisy\ if }\quad\frac{\mathrm{MDL}(\mathbf{s})}{\mathrm{len}(\mathbf{s})} < T\nonumber
\end{eqnarray}
If a sentence has no repeat patterns at all, then the length of the codebook should be $C\mathrm{len}(\mathbf{s})$, and $\mathrm{MDL}(\mathbf{s}) \ge C\mathrm{len}(\mathbf{s})$. Thus we choose $T=C$.

For the NLLB mined corpus, we remove pairs with laser score $< 1.06$ or language score $<0.95$ provided by LASER. Monolingual texts are cleaned using language scores only.

\begin{figure}[htb]
    \centering
    \includegraphics[width=\linewidth]{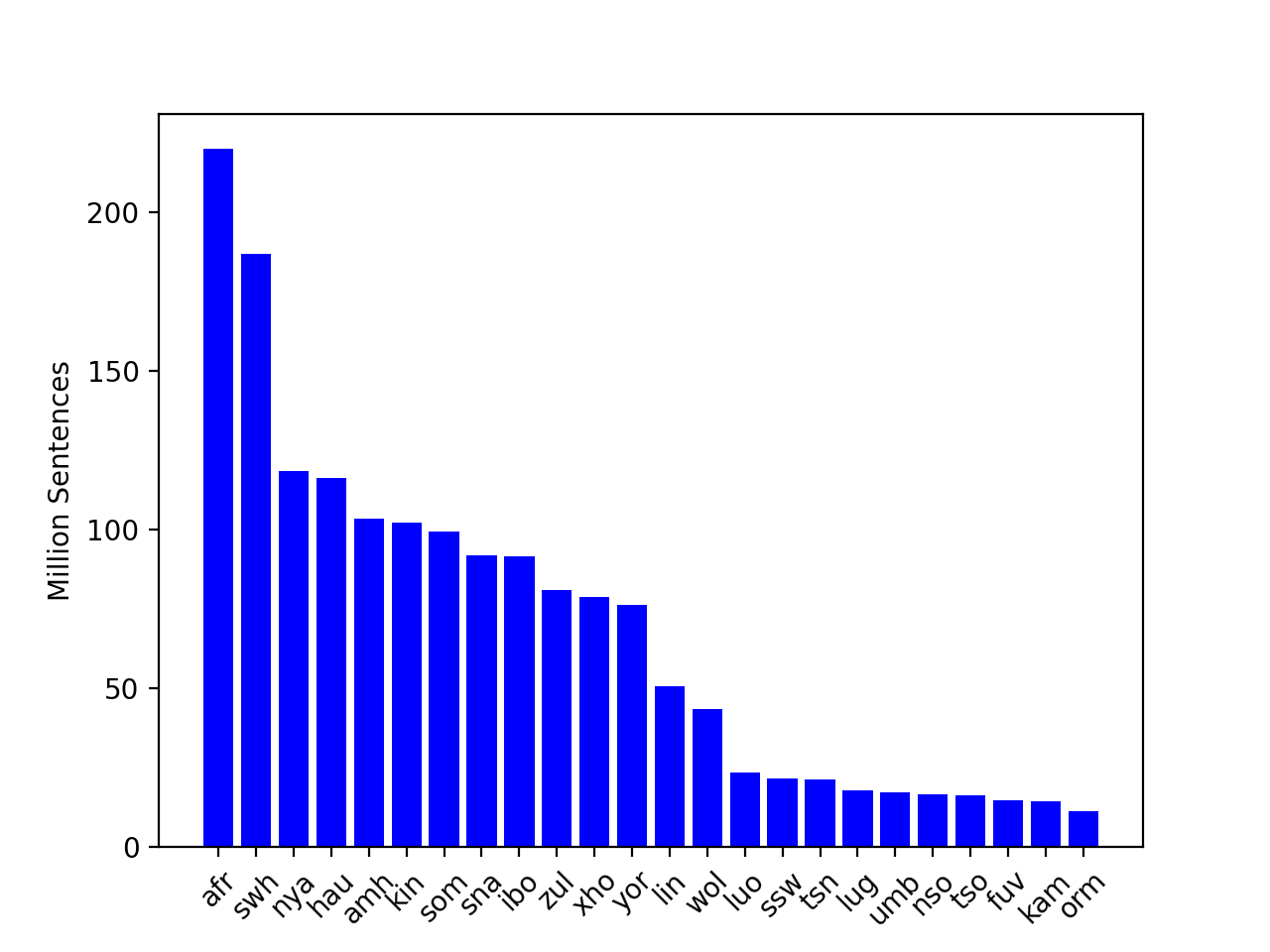}
    \caption{Number of sentences (in millions) in different African languages after data cleaning.}
    \label{fig:sentences}
\end{figure}

Table \ref{tab:datasize} and Figure \ref{fig:sentences} summarize the size of our training data after data cleaning and deduplication.

\begin{table}[ht]
    \centering
    \begin{tabular}{cc}
        Source & Sentence Pairs \\ \hline
        Constrained Track & $50.5M$ \\ 
        NLLB  & $29.1M$ \\ 
        Self Collected & $151.6M$ \\ 
        Back Translation& $1.41B$ \\ \hline
        Total & $1.64B$
    \end{tabular}
    \caption{Number of sentence pairs from different sources after data cleaning.}
    \label{tab:datasize}
\end{table}

\subsection{Preprocessing and Post Processing}

There are thousands of languages in the world, thus statically training a tokenizer on a predefined list of languages is not flexible for new languages. There are several studies on dynamic vocabulary for new language adaption, the general principle is to maximize the overlap with the old vocabulary. \cite{lakew-etal-2018-transfer,lakew-etal-2019-adapting}

We reuse the mRASP2 tokenizer, a unigram model trained on 150 languages using SentencePiece \cite{pan-etal-2021-contrastive}. To support new African languages, we train another tokenizer for new languages and merge it to the mRASP2 tokenizer. To ensure that the merged tokenizer produces the same segmentation for old languages, new words that can be made by joining two or more old words are removed and the rest new words' probabilities are scaled down.

We notice that the Yoruba text in FLORES200 has more accented characters than other corpora. 
According to NLLB team's report, the way FLORES200 marks the tone of vowels is similar to MAFAND dataset \cite{adelani-etal-2022-thousand}. Thus, we use the MAFAND data to train an accent model to post-process the translated sentences for $X\rightarrow$ Yoruba translation. It takes the Yoruba character sequence with accents removed as the input and outputs the accented characters. The structure of the model is a two-layer bidirectional LSTM having 50 hidden units in each layer. Correspondingly, we train another accent model using non-MAFAND datasets to preprocess source text for Yoruba$\rightarrow X$ translation.

\section{Model}

\subsection{Model Architecture}

Existing research works demonstrate that small models suffer from the underfitting problem for multilingual machine translation.  On the other hand, training and serving large models are expensive. Sometimes model parallelism or pipeline parallelism is necessary if it is impossible to run training on a single GPU due to memory constraints. And quantization is required to reduce the latency of inference. Our compromised model is a pre-layer norm transformer with $2.1B$ parameters which can be trained using $A100$ GPUs with $80G$ memory without parallelism. Details of the model are described in Table \ref{tab:model} 

\begin{table}[ht]
    \centering
    \begin{tabular}{c|c}
         Parameter & Value \\ \hline
         Encoder Layer &  $64$\\ 
        Decoder Layer &  $64$\\ 
        Hidden Size &  $1024$\\  
        FFN dimension &  $4096$\\  
        Max Length &  $512$\\  
        Shared Embedding &  Decoder input output \\
        Positional Embedding &  Learned  
    \end{tabular}
    \caption{Architecture of our transformer model}
    \label{tab:model}
\end{table}

\subsection{Language Tag}

There are two popular language tag strategies for multilingual MT: S-ENC-T-DEC which adds source language token to encoder input and target language token to decoder input \cite{JMLR:v22:20-1307,TACL2107,wu-etal-2021-language}, and T-ENC which adds target language token to encoder input \cite{yang-etal-2021-multilingual-machine,wu-etal-2021-language}. Our system uses T-ENC-T-DEC which adds the target language token to both encoder and decoder inputs. We did not use source language information for two reasons. First, most translation engines detect input languages automatically, which may introduce incorrect source language tokens. Second, a source sentence may be written in mixed languages. 


\section{Training and Optimization}

\subsection{Platform}

Our models are trained on $6$ machines each equipped with $8$ Nvidia $A100$ $80G$ GPUs. We use our internal version of ParaGen \footnote{https://github.com/bytedance/ParaGen} \cite{https://doi.org/10.48550/arxiv.2210.03405} , a self-developed text generation framework, to train the model. For back-translation, monolingual data are split and translated in parallel using $50$ Nvidia Tesla $V100$ GPUs.

To accelerate training, LightSeq is integrated. Unlike approaches that proposed alternative model structures to trade quality for speed, LightSeq used a series of GPU optimization techniques tailored to the specific computation flow and memory access
patterns of transformer models. It has been demonstrated $50\%$ to $250\%$ faster than Apex \footnote{https://github.com/NVIDIA/apex} on machine translation tasks. \cite{wang-etal-2021-lightseq,DBLP:journals/corr/abs-2110-05722}  Its inference speed is about $11.5$ sentences per second using a single Nvidia Tesla $V100$ GPU, which allows us to translate all monolingual texts within a month. 

As the training set's size exceeds the local disk's capacity, it is stored on a remote Hadoop file system. 

\subsection{Hyper-parameter Tuning}

We tune the hyperparameters using a hill climbing approach where each iteration searches along one direction with a different value in the hyperparameter space while keeping the others constant in order to converge to the locally optimal solution on the validation set. To search efficiently, we fix a small batch size and tune other parameters, then increase the batch size after the other parameters have been tuned. 

The final configuration is listed in Table \ref{tab:hyperparameters}. 

\begin{table}[ht]
    \centering
    \begin{tabular}{c|c}
        Hyperparameter & Value \\ \hline
        Initial Learning Rate  & $0.001$  \\ 
        Warmup Steps  & $1000$  \\ 
        Learning Rate Scheduler  & Inverse Square Root  \\
        Dropout Rate & $0.1$ \\
        Sampling Temperature & $5$ \\
        Label Smoothing & $0.1$ \\
        Optimizer & AdamW($0.9, 0.98$) \\
        Activation Function & ReLU \\
        Batch Size & $21M$ tokens  
    \end{tabular}       
    \caption{Hyperparameters for training.}
    \label{tab:hyperparameters}
\end{table}

\subsection{Streaming Data Shuffling}

Data Shuffling reduces the variance of mini-batches and lowers the risk of local optimum. However, it is challenging to shuffle a Terabyte-scale dataset dynamically. Our system uses multi-source streaming data based shuffling, which maintains a small in-memory buffer and a set of file pointers that point to random offsets of the training set. Each time a file pointer is selected randomly and loads the next sample to the buffer. A batch of samples is drawn from the buffer randomly once the buffer is full. This approach takes the advantage of data prefetching for sequential access in the Hadoop file system. The randomness of the sampling is controlled by the number of file pointers and the size of the buffer. In our experiments, we use about $5k$ file pointers and $300G$ host memory for the buffer.

To compare with global dynamic shuffling, we run a simulation experiment. We train a model until convergence, then shuffle the full dataset statically, and continue training on the shuffled data. Repeat shuffling until no significant change in loss or performance. For clarity, the original model is named as $M_0$, and the model trained with $i-th$ round of shuffled data is named as $M_i$.

Table \ref{tab:shuffle} shows the averaged per token loss of the last 100 training steps and averaged BLEU of $M_i$ on English $\leftrightarrow$ African language translations. We observed a slight improvement in the first round, but no significant change in the second round. This experiment suggests that our shuffling method combined with a limited number of static shuffling is a good approximation of global dynamic shuffling.

\begin{table}[ht]
    \centering
    \begin{tabular}{cccc}
         & $M_0$ & $M_1$  & $M_2$\\ \hline
        Averaged Loss  & $1.95$ & $1.91$ & $1.91$ \\ 
        Averaged BLEU  & $21.39$ & $21.48$ & $21.49$ \\ 
    \end{tabular}
    \caption{Simulation Experiment of global dynamic data shuffling: $M_0$ is the model trained on original training data. $M_i$ is the model trained on the $i-th$ round of statically shuffled data using $M_{i-1}$ as the initial point. The averaged training loss over the last 100 steps and averaged BLEU of English $\leftrightarrow$ African translations are reported.}
    \label{tab:shuffle}
\end{table}


\subsection{Small Dynamic Vocabulary vs Large Static Vocabulary}

Existing studies on vocabulary size do not reach a consensus. Large vocabularies often outperform small ones \cite{gowda-may-2020-finding}, but not always \cite{liao-etal-2021-back}

Our vocabulary has $100k$ words, smaller than most  of the other systems. Another difference is that our vocabulary is incrementally built for more than 150 languages, it may miss important words in new languages.

To understand the impact of vocabulary, we train another large unigram model with $200K$ words on the $26$ languages in this shared task.
Table \ref{tab:vocab} shows the performance with different vocabularies. It is obvious that the $100K$ vocabulary outperforms the $200K$ vocabulary, about $0.3$ improvement in BLEU on average. 

\begin{table}[ht]
    \centering
    \begin{tabular}{ccc}
        Vocabulary Size & Languages & BLEU \\ \hline
        $100k$ words & $173$ & $21.97$  \\ 
        $200k$ words & $26$ & $21.64$  \\ 
    \end{tabular}
    \caption{Average BLEU of English $\leftrightarrow$ African translations on the FLORES200 devtest set for the models with different vocabularies.}
    \label{tab:vocab}
\end{table}

\subsection{Pivot vs Direct}

As reported in Microsoft's work, pivot-based translation is more robust, especially for directions between low-resource languages since
corpora of $X\leftrightarrow Y$ are commonly sparser than $X\leftrightarrow $ English. \cite{yang-etal-2021-multilingual-machine} Therefore we use English as the pivot language for African-African translation. For French-African translation, the size of $X\leftrightarrow $ French data is comparable to $X\leftrightarrow $ English. Thus, we train a model for both English and French and choose the better one during inference time.

\subsection{Model Averaging}

As suggested by other works, model averaging is a simple trick that could significantly improve the performance without changing the model structure or slowing the inference speed. The only cost is the external disk spaces to save intermediate checkpoints, which is trivial compared with GPU and memory costs. 

We save the checkpoints every 100 updates of gradients and average the last $K$ checkpoints. By enumerating $K$ from $1$ to $20$, we find that $K=10$ is large enough to capture most of the gains. 

\section{Results}

\subsection{System Tuning}
We tune our model on the FLORES200 devtest dataset, starting with a base model trained on the official data for the constrained track. Then we add more datasets and apply the  optimization described above to boost performance. Table \ref{tab:tune} reports the averaged BLEU over 56 directions including 24 African languages from and to English and 4 African languages from and to French.

\begin{table}[htb]
    \centering
    \begin{tabular}{lc}
    Model Description & BLEU \\\hline
         Base model   & $16.92$ \\
         + NLLB and self-collected data  & $18.89$ \\
         + Data cleaning  & $19.64$ \\
         + Back-translation data  & $22.85$ \\
         + $X\rightarrow$ English $\rightarrow$ French & $22.95$ \\
         \ \ \ \  + French  $\rightarrow$  English $\rightarrow X$ {}\textsuperscript{\textdagger} & $22.90$ \\
         + Yoruba Accent for $X \rightarrow$ Yoruba & $23.20$ \\
          \ \ \ \ + Yoruba Accent for Yoruba $\rightarrow X$ {}\textsuperscript{\textdagger} & $23.17$ \\
         + Model Averaging & $23.35$  
    \end{tabular}
    \caption{System tuning on FLORES200 devtest set, averaged BLEU over 56 directions is reported. Superscript {}\textsuperscript{\textdagger} means the modification is not included in the final submission.}
    \label{tab:tune}
\end{table}

We can see that the amount of training data is proportional to the performance of the model, especially when back-translation data is added. For some very low resource languages such as Wolof, back-translation improves Wolof $\rightarrow$ English from $11.1$ to $19.3$, and English $\rightarrow$ Wolof from $4.17$ to $7.07$.

Another observation is that pivot translation outperforms direct translation for $X \rightarrow$ French directions, but underperforms for French $\rightarrow X$, which indicates that the final step in pivot translation dominates the overall performance.

The impact of Yoruba accent models also shows mixed results. There is  a significant improvement for $X\rightarrow$ Yoruba translation, but a little damage to  Yoruba$\rightarrow X$ translation. One possible reason is that the non-MAFAND dataset has multiple sources with different accent annotation standards, making the accent model confused. Therefore we only apply post-processing for $X\rightarrow$ Yoruba translations.

\subsection{Final Result}

Official evaluation metrics include BLEU, sentence-piece BLEU (spBLEU) score, and chrF++. Table \ref{tab:res} shows the results of our primary submission on FLORES200 dev, FLORES200 devtest set, and hidden test sets respectively. The sentence-piece model for calculating spBLEU is SPM-200 provided by Meta AI \footnote{https://github.com/facebookresearch/fairseq/tree/nllb}

\begin{table}[ht]
    \centering
    \begin{tabular}{cccc}
        Dataset & BLEU & spBLEU & chrF++ \\ \hline
        FLORES dev  & $17.41$ &  $21.70$ & $42.01$  \\ 
        FLORES devtest  & $17.43$ & $21.71$ & $41.99$ \\ 
        Official test  & $17.30$ & $21.90$ & $41.87$ \\ 	
    \end{tabular}
    \caption{Results of our primary submission on FLORES200 dev, FLORES200 devtest and official test datasets respectively. Metrics are averaged over 100 language pairs.}
    \label{tab:res}
\end{table}

\section{Conclusion}

This paper presents our system for the WMT22 shared task on Multilingual Machine Translation for African Languages. We focus on data collection, augmentation, and cleaning. Due to the limited time, we did not try modeling tricks such as reranking and ensemble. Our finding is that the amount of data is crucial for translation quality, especially monolingual data in low-resource languages. 

\section*{Acknowledgements}
We thank Ying Xiong and Yang Wei for building the LightSeq package for this submission. We also thank the GMU-eval team for their effort to make our system work on the evaluation platform.

\bibliography{anthology,custom}
\bibliographystyle{acl_natbib}

\end{document}